# A Multilayered Approach to Classifying Customer Responsiveness and Credit Risk

Ayomide Afolabi, Ebere Ogburu and Symon Kimitei
School of Data Science, Kennesaw State University, Kennesaw, USA
Email: aafolab4@students.kennesaw.edu ; eogburu@students.kennesaw.edu;  skimitei@kennesaw.edu

## ABSTRACT

This study evaluates the performance of various classifiers in three distinct models: response, risk, and response-risk, concerning credit card mail campaigns and default prediction. In the response model, the Extra Trees classifier demonstrates the highest recall level (79.1%), emphasizing its effectiveness in identifying potential responders to targeted credit card offers. Conversely, in the risk model, the Random Forest classifier exhibits remarkable specificity of 84.1%, crucial for identifying customers least likely to default. Furthermore, in the multi-class response-risk model, the Random Forest classifier achieves the highest accuracy (83.2%), indicating its efficacy in discerning both potential responders to credit card mail campaign and low-risk credit card users. In this study, we optimized various performance metrics to solve a specific credit risk and mail responsiveness business problem.

## I. INTRODUCTION

Credit card users' payment delinquency or default contributes to the rising credit card debt and charge-offs, which poses a financial risk that can adversely affect the profitability of banks and other credit card issuing companies. As of December 2023, United States' consumer revolving credit card debt stands at an all-time peak of $1.3 trillion [1]. Meanwhile, the delinquency rate past thirty days or more stands at 2.98 percent, with a charge-off rate of 3.79 percent [2] for average loans removed from the books and charged against loss reserves as of the end of the third quarter of 2023. Considering that this charge-off could potentially impact the profitability of these companies, it has become necessary for credit card issuers to devise various strategies to shield themselves from exposure to such risk resulting from credit card delinquency or default by identifying potential credit users' defaulters early.

In recent times, the utilization of machine learning and statistical models to predict credit card payment defaulters has been on the rise due to their ability to achieve high accuracy and recall, and these models have been employed in different studies [4,5,6], which has proven helpful in identifying customers who are more likely to be delinquent on payments which might lead to a charge-off. However, to the best of our knowledge, no studies have fully explored the use of machine learning models to identify potential customers who are more likely to respond to a credit card offer and while also not being delinquent in a targeted marketing strategy.

In this study, we hope to explore different machine learning models that can identify potential credit users who are more likely to respond to a credit card offer and minimize any chances of delinquency from customers in a targeted marketing strategy.

## II. LITERATURE REVIEW

As highlighted by Siddiqi [13], assessing a customer's creditworthiness typically entails developing a statistical model based on their historical data, applying the model to prospective customers to gauge their creditworthiness, and continually monitoring its effectiveness using established business metrics. Utilizing a classification model like logistic regression, as noted by Abdou and Pointon [14], facilitates the aggregation of customer data to predict repayment behavior or profitability accurately. To predict credit card defaulters, Venkatesh *et al.* [3] employed the use of traditional machine learning algorithms including Naive Bayes and Random Forest, while Fan [7] compared the predictive performance of support vector machines to decision trees.

To further improve the performance of these traditional machine learning models, Ebiaredoh-Mienye *et al.* [6] used features extracted from stacked sparse autoencoders as inputs for these models, and Almajid [12] addressed data imbalance issues usually associated with credit card data through a combination of one-hot encoding and Synthetic Minority Over-Sampling Technique (SMOTE). In exploring the use of deep learning models, Sun *et al.* [4] investigated the efficacy of deep learning algorithms in contrast to traditional machine learning methods for developing a predictive system to aid credit issuers in modeling credit card delinquency risk. Additionally, Kim *et al* [5] proposed variational deep embedding with sequence based on deep neural networks which outperforms other baseline models to predict future sequence delinquency of customers.

Ensemble techniques have been extensively explored in the classification of credit card users. Studies such as Lee et al [8] focused on employing algorithms like LightGBM, XGBoost, and CatBoost for this purpose. Similarly, Lawi et al [9] introduced a least squares SVM ensemble method, while Zhao et al [10] and Gao [11] delved into the effectiveness of CatBoost and XGBoost-LSTM, respectively.

## III. MATERIALS AND METHODS

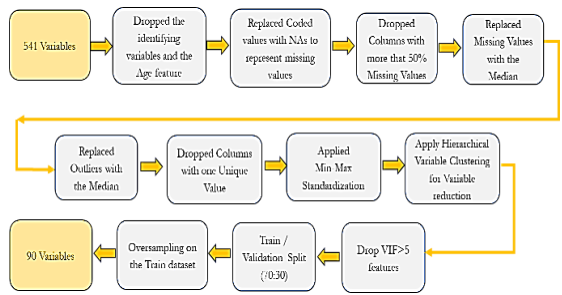

Figure 1: Data Preprocessing Flowchart

The Atlanticus dataset contains several coded values which represent missing digit values and can be identified uniquely since they fall in the range of 2 to 9 and 92 to 99999. These numbers were flagged and replaced with NAs. Given that all predictor variables in the dataset are continuous, straightforward imputation methods such as median imputation were utilized to handle the missing values. Variables with only one unique value and those with over 50% missing values were removed from the dataset to maintain data integrity. This process reduced the number of features from 541 to 465.

Regarding the label variable, the missing goodbad category is defined as customers who did not open a tradeline and did not respond to the direct mail campaign. The missing goodbad is assigned a value of 2 to appropriately represent the individuals who never opened a credit card. It's important to note that the dataset that we analyze consists of customers who responded to the direct mail campaign and are categorized more formally below to ensure a comprehensive analysis of the classification as follows.

$$y_i = \begin{cases} 0 & \text{if individual opened card and not delinquent} \\ 1 & \text{if individual opened card and is delinquent} \\ 2 & \text{if individual didn't open a card} \end{cases} \quad \ldots\ldots\ldots (1)$$

In the dataset, 15.6% of all observations have a class of 0, 9.2% of all observations have a class of 1 and 75.2% of all the observations have a class of 2.

Moreover, values deviating more than 3 standard deviations from the mean were replaced with the median value of each feature to mitigate the influence of outliers. Additionally, to ensure that the varying scales of different features is minimized, min-max scaling technique is utilized. The normalization technique ensures different feature values are shifted and rescaled to fall between 0 and 1.

To select appropriate variables from the normalized variables, clustering is performed. Using the hierarchical clustering technique, groups of similar features are identified and then visualized in a dendrogram. A variable in a particular cluster is selected as a cluster representative variable based on its high R-squared value within its own cluster which further reduces our variables set.

From the set of the reduced variables, to detect and address the issue of multicollinearity that might exist, we utilized the variance inflation factor method. Multicollinearity exists when there are highly correlated predictor variables which cause the standard errors to be inflated and leads to unreliable coefficient estimates. The predictor variables with a variance inflation factor greater than 5 were dropped from the dataset as they pose a cause for concern. Handling this provided the desired variables needed for the modeling process.

Three models are utilized to address the business problem: a response model, risk model, and response-risk model. To develop the response model, customers who were delinquent and non-delinquent are merged to generate a new class 1 representing those who responded to the mail and customers who did not open a card are assigned a new class 0 to represent those who did not respond to the mail. The response-risk model is used to further assess the efficacy of the other two binary classification models.

Prior to building these machine learning models, a train-validation split of the dataset is performed, with 70% for training and 30% for the validation to optimize the model's performance.

The analysis of the dataset revealed a class imbalance issue with 15.56% of the observations being individuals who opened a card without delinquency, 9.24% are those who opened a card with delinquency, and 75.2% being those who never opened a card. As evidenced in the distribution, skewed class proportions exist. Customers who did not open a card make up a sizable proportion of the dataset. When this problem arises, standard classifiers tend to become biased to the large classes over the small ones and this often leads to false performance metrics. To handle this problem, we employed the Adaptive Synthetic Sampling Approach for Imbalanced Learning (ADASYN) python library to oversample by generating synthetic samples for minority instances that are difficult to classify using the training dataset. It is pivotal to perform oversampling on the training dataset to eliminate instances of data leakage and overfitting.

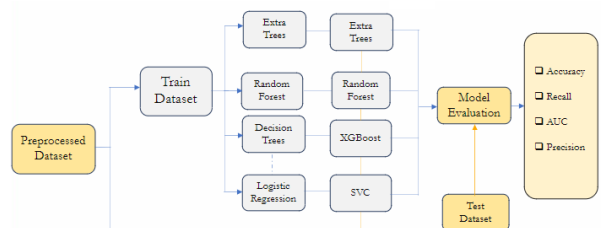

Figure 2: Model Building Flowchart

Consecutively, the training dataset is utilized to fit several machine learning algorithms on all three models: response, risk, and response-risk. Through this process, the best-performing models are selected, and these models are subsequently applied to the validation dataset. To ensure the models are optimized for our objectives, several performance metrics are employed. Accuracy measures the proportion of correct predictions out of the total predictions made. Precision quantifies the fraction of true positive predictions among all positive predictions, whereas recall calculates the fraction of true positive predictions out of the total actual positive instances. The Area Under the Receiver Operating Characteristic Curve (AUC) is another important metric used, as it measures the model's ability to distinguish between positive and negative classes across various threshold settings. Specificity is also used to evaluate the risk model's ability to correctly identify the negative instances (no default).

An integral procedure applied to provide these stable and robust results was to tune the hyperparameters. Hyperparameter tuning is a crucial step in optimizing machine learning models, as it enables the adjustment of model parameters to achieve better performance on specific metrics. By systematically exploring the hyperparameter space, we were able to identify the optimal configurations that maximized the trade-off between various performance measures, such as accuracy, precision, recall, AUC, etc.

## IV. RESULT AND DISCUSSION

### i.) The Response Model

Table 1 below highlights the performance of the three best classifiers we explored for creating the response model, with the Extra Trees classifier showcasing the highest recall value of 79.1%. Notably, recall stands out as a crucial metric for identifying potential responders to targeted credit card mail offers. This also signifies the proportion of customers likely to positively respond to the mail offer, making it a pivotal measure for the mail campaign success.

| Response Model | | | | | | | | |
|---|---|---|---|---|---|---|---|---|
| | Accuracy | | Precision | | Recall | | Auc | |
| Models | Train | Validation | Train | Validation | Train | Validation | Train | Validation |
| Random Forest Classifier | 0.663 | 0.588 | 0.635 | 0.347 | 0.788 | 0.745 | 0.718 | 0.686 |
| XGBoost | 0.982 | 0.814 | 0.998 | 0.693 | 0.965 | 0.455 | 0.999 | 0.784 |
| Extra Trees Classifier | 0.641 | 0.561 | 0.612 | 0.337 | 0.794 | 0.791 | 0.694 | 0.693 |

Table 1: Performance Metric of the Top Three Response Classifiers

| Response Model Tuning Hyperparameter | | |
|---|---|---|
| Models | Maximum Depth | Number of Estimators |
| Random Forest Classifier | 1 | 60 |
| XGBoost | 16 | 265 |
| Extra Trees Classifier | 1 | 280 |

Table 2: Response Model Hyperparameters

Based on the data in Figure 1, we can identify some valuable insights into the performance of our model in predicting customer responses to our targeted mail campaign. Firstly, the figure reveals that the model successfully identified 58,297 True Positives (TP), indicating instances where it correctly predicted customers who would respond to the mail campaign. This outcome is highly desirable, as it represents potential opportunities for engagement and conversion. Similarly, the model achieved 38,842 True Negatives (TN), accurately identifying customers who would not respond to the campaign. This outcome is also favorable, as it ensures that resources are not wasted on uninterested parties.

However, there were 114,859 False Positives (FP), where the model incorrectly classified customers as responders when they were not. These instances can be considered false alarms and represent a missed opportunity to allocate resources more effectively. Most concerningly, the model predicted 15,440 False Negatives (FN) as customers who would respond to the mail campaign when they should be labeled as non-responders. This outcome is the least desirable, as it represents lost opportunities for engagement and conversion.

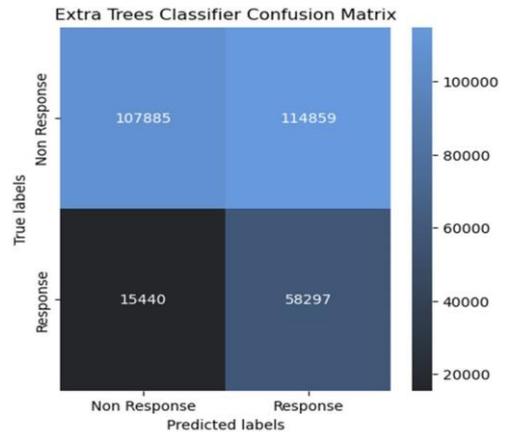

Figure 1: Confusion Matrix of the Response Extra Trees Classifier

In Figure 2 presented below, the ROC plot demonstrates the relationship between sensitivity and specificity across various decision threshold values. The Area Under Curve (AUC) serves as a comprehensive measure of the model's diagnostic accuracy. An AUC of 0.5 indicates a performance equivalent to random chance, while an AUC of 1.0 signifies perfect accuracy. Despite XGBoost exhibiting the highest AUC of 0.789, it lacks reliability since the primary objective of the experiment is to identify the model with the highest recall value, which, in this case, is achieved by the Random Forest model as indicated in table 1.

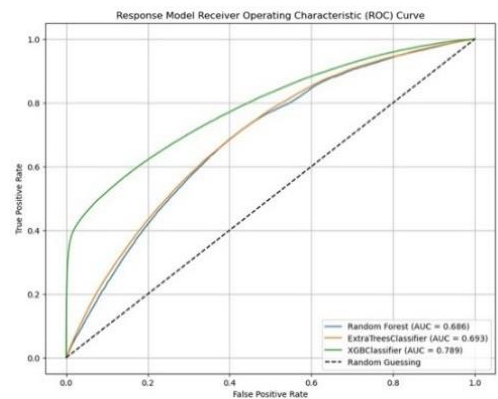

Figure 2: ROC Curve for the Response Extra Trees Classifier

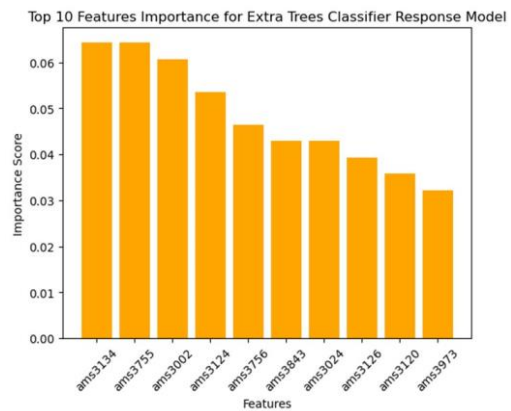

Figure 3: Top 10 Important Response Model Features

### ii.) The Risk Model

Table 3 highlights the performance of the three best classifiers we explored for the Risk model, with the Random Forest classifier showcasing the highest specificity value of 0.841. Notably, specificity stands out as a crucial metric for identifying customers who are most likely not to default during their credit usage. The result is based on the optimized model after hyperparameter tuning focusing on maximizing the specificity metric. The optimal hyperparameters for the Random Forest classifier were found to be a maximum depth of 1 and 280 estimators as depicted in table 4.

| Risk Model | | | | | | | | |
|---|---|---|---|---|---|---|---|---|
| | Accuracy | | Specificity | | Recall | | Auc | |
| Models | Train | Validation | Train | Validation | Train | Validation | Train | Validation |
| Random Forest Classifier | 0.995 | 0.735 | 0.994 | 0.841 | 0.995 | 0.557 | 1.000 | 0.780 |
| XGBoost | 0.995 | 0.731 | 0.994 | 0.836 | 0.995 | 0.555 | 1.000 | 0.779 |
| Extra Trees Classifier | 0.991 | 0.732 | 0.989 | 0.826 | 0.994 | 0.575 | 1.000 | 0.779 |

Table 3: Performance Metrics of the Top Three Risk Classifiers

| Response Model Tuning Hyperparameter | | |
|---|---|---|
| Models | Maximum Depth | Number of Estimators |
| Random Forest Classifier | 1 | 60 |
| XGBoost | 16 | 265 |
| Extra Trees Classifier | 1 | 280 |

Table 4: Risk Model Hyperparameters

Figure 4 illustrates that there were 15,275 instances of True Positives (TP), which represents the number of customers who are predicted as those who would default. Conversely, the figure also indicates 38,842 True Negatives (TN), reflecting the correct identification of non-defaulting customers. However, the model generated 12214 False Negatives (FN), thus categorizing customers as defaulters when they were not. These cases represent false alarms, as the customers are, in fact, non-defaulters.

Furthermore, there were 7406 False Positives (FP), which represents the number of customers who are predicted as those who would default but are incorrectly classified as non-defaulters by the Random Forest model. This outcome is particularly undesirable, as it signifies missed opportunities to identify risky customers accurately.

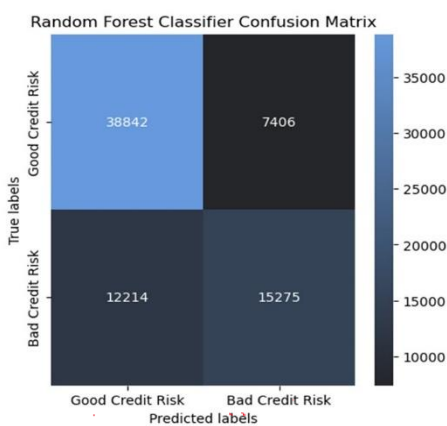

Figure 4: Confusion Matrix for the Risk Model Random Forest

As depicted in figure 5, all the three models have the same AUC of approximately 0.779. This indicates that the metric lack reliability. Since the primary objective of the experiment is to identify the model with the highest specificity value, we selected the Random Forest model as indicated in table 3. For instance, a specificity value of .84 means that from a total of 100 actual non-defaulters, the random forest model predicted 84 of them correctly.

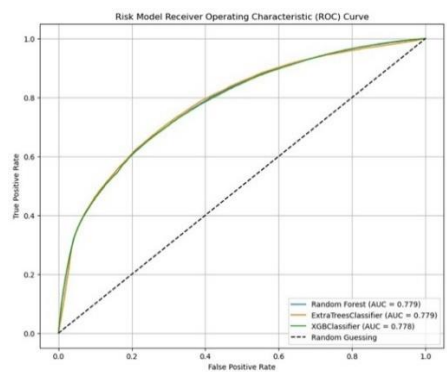

Figure 5: Risk Model Random Forest ROC curve

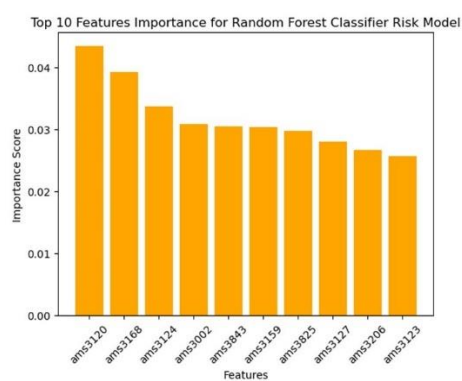

Figure 6: Top 10 Important Risk Model Features

Ultimately, profitability is determined by quantifying the credit risk associated with the risk model. When a customer is classified as bad credit risk a charge of -$600 is assigned. While a positive charge of $200 is assigned when a good credit risk customer is selected. The profitability of using a machine learning model is computed by calculating the dot product between the profitability scoring matrix and the confusion matrix. As seen in table 5, the profit of using XGBoost, Extra Trees Classifier and Random Forest Classifier are $5,122,400, $4,709,400, and $5,309,400 respectively. From this result, the Random Forest Classifier provides the highest profit.

| Risk Model Profitability | | | |
|---|---|---|---|
| | XGBoost | Extra Trees Classifier | Random Forest Classifier |
| Profit | $5,122,400 | $4,709,400 | $5,309,400 |

Table 5: Risk Models Maximum Profit Attained

### iii.) The Multi-Class Response-Risk Model

Table 6 below highlights the performance of the three best classifiers we explored for creating the Response-Risk model. The Random Forest classifier had the highest accuracy value of 0.823 and AUC of 0.786. Accuracy is a useful metric for multiclass problems because it provides an overall measure of the model's performance across all classes. These were selected as the best metrics for identifying the proportion of customers who are most likely to respond to the targeted market mail campaign and those who are most likely to not default on the credit card. The hyperparameters for the Response-Risk Model were tuned, focusing on maximizing accuracy. For best model, which is the Random Forest Classifier, the optimal hyperparameters were a maximum depth of 42 and 337 estimators.

| Response-Risk Model | | | | | | | | |
|---|---|---|---|---|---|---|---|---|
| | Accuracy | | Precision | | Recall | | Auc | |
| Models | Train | Validation | Train | Validation | Train | Validation | Train | Validation |
| Random Forest Classifier | 0.996 | 0.823 | 0.996 | 0.763 | 0.996 | 0.559 | 1.000 | 0.786 |
| XGBoost | 0.996 | 0.823 | 0.994 | 0.766 | 0.996 | 0.557 | 1.000 | 0.783 |
| Extra Trees Classifier | 0.993 | 0.815 | 0.993 | 0.720 | 0.993 | 0.563 | 1.000 | 0.780 |

Table 6: Performance Metrics of the Top Three Response-Risk Classifiers

| Response-Risk Model Tuning Hyperparameter | | |
|---|---|---|
| Models | Maximum Depth | Number of Estimators |
| Random Forest Classifier | 42 | 337 |
| XGBoost | 24 | 33 |
| Extra Trees Classifier | 47 | 310 |

Table 7: Model-Risk Model Hyperparameters

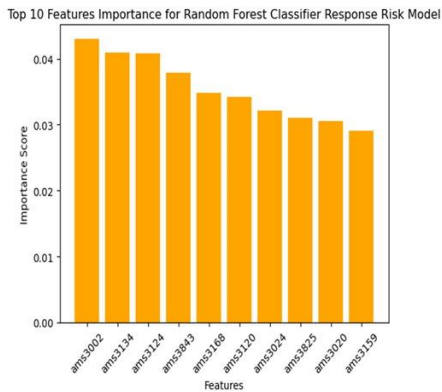

Figure 7: Top 10 Important Risk Model Features

The top 10 most important features are shown in figure 3, figure 6 and figure 7. Interestingly, two features appear to be ranked highly in all three models. These features are ams3002 and ams3124 which represent the number of inquiries made within 24 months and 6 months respectively. It would be interesting to see if these features become highly ranked in ensemble models. These features demonstrate high predictive power and robustness across the three modeling approaches for the business problem.

## V. CONCLUSION

In conclusion, the Extra Trees classifier stands out as the optimal choice for accurately predicting customers likely to respond to a targeted mail marketing campaign. In the response-risk multiclass situation, the Random Forest classifier achieved the best performance.

Our results suggest that when identifying booked customers who are most likely not to default, a machine learning model optimized for true negatives is best. As evidenced by the specificity metric, the Random Forest classifier excels at identifying customers most likely not to default (good credit risk). Consequently, this classifier offers the highest profitability.

This project underscores the strength of machine learning methodologies in pinpointing potential credit users primed to respond to a credit card offer, while concurrently mitigating the risk of delinquency within a targeted marketing approach.